\documentclass[10pt,a4paper,twocolumn]{article}

\usepackage[T1]{fontenc}
\usepackage[utf8]{inputenc}
\usepackage{lmodern}
\usepackage[a4paper,margin=1.8cm,columnsep=0.8cm]{geometry}
\usepackage{times}
\usepackage{authblk}
\usepackage{titlesec}
\usepackage{parskip}
\usepackage{url}

\titleformat{\section}
  {\normalfont\large\bfseries}
  {\thesection.}{0.5em}{}

\title{\vspace{-1em}\LARGE\bfseries What Does It Mean for a Medical AI System to Be Right?}

\author[1]{Antony M. Gitau}
\affil[1]{University of South-Eastern Norway}
\date{}

\begin{document}

\twocolumn[
\maketitle
\vspace{-1.2em}
\begin{center}
\end{center}
\vspace{1em}
]

\section{Abstract}

This paper examines what it means for a medical AI system to be 'right' by grounding the question in a specific clinical context: the automatic classification of plasma cells in digitized bone marrow smears for the diagnosis of multiple myeloma. Drawing on philosophy of science and research ethics, the paper argues that correctness in medical AI is not a singular property reducible to benchmark performance, but a multi-dimensional concept involving the availability of expertly labeled medical datasets, the explainability and interpretability of model outputs, the clinical meaningfulness of evaluation metrics, and the distribution of accountability in human-AI workflows. As such, the paper develops this argument through four interrelated themes: the instability of ground truth labels, the opacity of overconfident AI, the inadequacy of standard clinical metrics, and the risk of automation bias in time-pressured clinical settings. 

\section{Introduction}

The promise of AI in medical diagnostics has never been more loudly proclaimed. For example, classification models now approach or exceed specialist performance on narrow benchmarks in radiology, pathology, and dermatology. When a model is said to perform well, for a task such as classification, the area under the receiver operating characteristic curve (AUC-ROC) and F1-score are the standard metrics. But these metrics presuppose a correct expert-generated ground truth of the classification dataset against which a model's predictions can be compared. In medical imaging, researchers have uncovered incorrect labels, referred to as label noise, in existing public datasets \cite{northcutt2021confident}. This means that when models are developed using such data, errors are likely to occur.

Automatic classification of plasma cells in digitized bone marrow smears for the diagnosis of multiple myeloma is an example of a medical imaging task, which is one of my PhD research focus areas. Bone marrow morphology is one of the most inter-observer variable domains in diagnostic pathology. Plasma cells exist on a morphological continuum from normal to highly abnormal; their percentage in the marrow is a key diagnostic criterion for myeloma, yet what counts as a plasma cell in a borderline case is a judgment call that experienced hematologists can and do resolve differently, reflecting known inter-observer variability in diagnostic pathology \cite{elmore2015diagnostic}. 

The argument proceeds in four steps. First, we will examine the nature of ground truth in bone marrow morphology datasets. Then, consider what follows ethically for model design. Specifically, whether confident predictions are always appropriate, or whether calibrated uncertainty is an epistemic and ethical requirement. Third, discuss standard evaluation metrics and show how they can mask clinically dangerous failure modes in a rare-cell detection task. Finally, address the distribution of responsibility when AI-assisted diagnosis goes wrong, focusing on the problem of automation bias in high-stakes, time-pressured settings.
The paper does not claim that correctness in medical AI is impossible to achieve or assess. Rather, it argues that correctness is plural and context-dependent, and that a system can only be considered right if it is appropriately uncertain in the face of ambiguity, that is, epistemically humble, clinically useful, and embedded in human oversight structures that preserve genuine accountability.

\section{Instability of ground truth labels}

A standard supervised machine learning pipeline in a medical imaging task contains an image and its accompanying labels. With a dataset assembled and annotations collected, a model is trained to predict those labels, and performance is reported against a held-out test set. The implicit ontology here is that there exists a true class membership for each data point, which annotators assigned, and the model learns to recover. This assumption is defensible. 

To explain where the instability of ground truth stems from, we will explore a bone marrow case example. Plasma cells in a bone marrow aspirate smear vary enormously in appearance. The boundary between reactive plasmacytosis, a benign condition, and frank myeloma is defined partly by plasma cell percentage, with the World Health Organisation (WHO) 2022 criteria requiring at least 10\% clonal plasma cells for a myeloma diagnosis \cite{li2022who5th}. Counting and classifying these cells under a microscope is a task where inter-observer variability among trained pathologists can be clinically significant. Studies have documented meaningful disagreement rates between expert readers on bone marrow preparations sufficient to alter diagnosis in a non-trivial proportion of cases, and the same variability extends to digitized smear analysis.

Although better annotation solutions are being developed to solve this problem, we can also further examine the nature of diagnostic categories. Ludwig Fleck's writings in 1935 described how scientific facts are produced within 'thought collectives', communities of practice that share implicit standards, instruments, and conceptual frameworks \cite{fleck1979genesis}. A diagnosis is not a neutral reading of nature; it is a judgment made within a professional tradition, shaped by available treatment options, institutional protocols, and the specific training of the observer. When two hematologists disagree on whether a cell is a plasma cell, neither is simply wrong. They are applying slightly different versions of a collectively maintained standard that has never been fully codified.

For AI development, this has concrete implications. When labels are obtained from a single annotator, the model learns that annotator's particular version of the standard. When labels are aggregated using majority vote, disagreement is suppressed rather than represented. The fact that a cell was borderline is erased, replaced by a definitive label that conceals the uncertainty from which it emerged. When datasets are merged across institutions, divergent local conventions are collapsed into a single label set without any mechanism for tracking the underlying variation. In each case, the ground truth against which model performance is measured is not a stable fact of the world but a contingent social artifact. Reporting test-set accuracy of 94\% against such a ground truth says something real, but it does not say that the model's outputs correspond to clinical reality 94\% of the time.

A more epistemically honest approach would treat inter-annotator disagreement as a signal rather than noise. Soft labeling, which represents each training instance as a probability distribution over possible classes weighted by annotator confidence, has been proposed in medical image analysis as a way to preserve and learn from this information. At minimum, reporting inter-annotator agreement statistics (such as Cohen's kappa) as part of dataset documentation should be treated not as optional metadata but as a prerequisite for interpreting any downstream performance claim. The EU AI Act (2024), which classifies diagnostic AI as high-risk and requires high-quality datasets with documented data governance \cite{eu2024aiact}, implicitly demands this, although it does not specify what quality means for inherently ambiguous annotations,  a gap that domain-specific standards must fill.

To conclude this theme, correctness, in this first sense, is argued as not correspondence with mind-independent reality but coherence with a professional standard that is itself subject to interpretation and historical revision. This does not make AI-assisted plasma cell classification impossible; it means that claims about its correctness must be understood within, not independently of, that social and epistemic context.

\section{Opacity of overconfidence}

A second dimension of correctness concerns models' behaviour in the face of ambiguity. Standard classification architectures output a probability distribution over classes, and in practice, the highest-probability class is taken as the prediction. The architecture is capable of expressing uncertainty; a near-uniform distribution signals low confidence, but in deployment, this information is often discarded in favor of a single binary decision. The model commits to an answer, and the answer is treated as the model's output.

This compulsory confidence can be problematic in high-stakes medical settings. Consider a borderline cell from a bone marrow smear where a model assigns 52\% probability to 'plasma cell' and 48\% to 'precursor B-cell.' A confident prediction of 'plasma cell' in this case is not simply a classification; it is a claim that may influence whether a patient is diagnosed with multiple myeloma and begins a course of demanding systemic chemotherapy. The model's internal uncertainty, which is clinically highly relevant, has been erased by the convention of reporting only the output with the higher probability.

The alternative is calibrated uncertainty, which means ensuring that a model's stated confidence levels correspond to its actual accuracy rates \cite{guo2017calibration}. A model is well-calibrated if, across all instances where it assigns 70\% confidence to class A, it is correct approximately 70\% of the time. Research in medical imaging has consistently shown that modern deep neural networks tend towards overconfidence. They assign high probabilities to predictions that are wrong at non-trivial rates \cite{nixon2019measuring}. This is particularly dangerous in a safety-critical application where overconfidence cannot be detected from the output alone.

Beyond calibration, there is a compelling argument for selective prediction: the ability of a model to abstain from classification when its confidence falls below a threshold, flagging ambiguous cases for expert review \cite{secondopinion2021}. In the bone marrow context, a model that declines to classify morphologically unusual cells is making an epistemically honest and clinically responsible choice. This connects directly to the human-in-the-loop architecture that is central to my PhD work, where the AI development is intended to support rather than supplant pathologist judgment.

This position is also aligned with current transparency requirements. The WHO's ethics principles for AI in health include transparency as a core requirement, specifically the need for AI systems to be explainable and interpretable in ways clinicians can act on \cite{who2024ai}. DECIDE-AI reporting guidelines require documentation of how AI uncertainty is communicated to end users \cite{decideai2022}. Explainability methods, such as attention maps and saliency visualisations, allow visual inspection of the features driving a prediction. But explainability cannot compensate for a model that misrepresents its own uncertainty. Knowing what the model looks at is not the same as knowing how much to trust what it says.

There is a deeper connection here to Popperian philosophy of science. Popper argued that scientific claims earn their epistemic standing by being falsifiable \cite{popper1959logic}. A medical AI system that always outputs a confident prediction, regardless of input ambiguity, is in a relevant sense unfalsifiable from the outside. It always produces an answer, so no datum could register as a failure of the system's own stated confidence. A model that expresses calibrated uncertainty, or that abstains on difficult cases, makes weaker claims in situations where weaker claims are epistemically warranted, creating opportunities for its errors to be detected, reported, and corrected. This is a practical requirement for the iterative learning and improvement that safe clinical AI deployment demands.

\section{Inadequacy of metrics}

A third dimension of correctness concerns how performance is measured. Evaluation metrics do not neutrally reflect model quality; they actively shape what it means for a model to be right, and different metrics can tell profoundly different stories about the same system.

In bone marrow smear analysis for multiple myeloma, plasma cells are the diagnostically critical cell type, but they may constitute only 1 to 15 percent of all cells in a smear at the time of initial assessment, or even less in early-stage disease. This creates a class imbalance problem with direct clinical consequences. A model trained to maximise overall classification accuracy can achieve deceptively high scores by correctly classifying the abundant non-plasma cells while performing poorly on the rare plasma cells that determine the diagnosis. Accuracy, in this context, is not merely incomplete; it is potentially misleading, capable of allowing a clinically dangerous model to pass standard evaluation.

More appropriate metrics for heavily imbalanced problems include precision, recall, and the area under the precision-recall curve (AUPRC), which is more informative than AUC-ROC when class imbalance is severe. In clinical terms, sensitivity over the plasma cell class determines what proportion of plasma cells are detected, governing the risk of underestimating plasma cell burden and missing a myeloma diagnosis. The clinical stakes of sensitivity and specificity errors differ substantially and context-dependently. In screening, high sensitivity takes priority, while when confirming an existing clinical suspicion, the trade-off may be calibrated differently. Standard aggregate metrics do not encode this clinical context, a limitation that can only be addressed by reporting disaggregated performance tied to clinically defined thresholds.

Even well-chosen aggregate metrics have residual limitations. They evaluate performance averaged across a test set, and averaging can conceal systematic failure in particular morphological subtypes. A model might perform well on typical plasma cell morphologies while failing on plasmablastic variants, most common in high-grade disease, precisely the cases where correct classification matters most for treatment planning. Such systematic errors may be invisible in aggregate reporting but detectable through stratified analysis by morphological subtype or clinical stage.

The CONSORT-AI and SPIRIT-AI reporting guidelines respond to some of these concerns by requiring subgroup performance analyses and transparent reporting of failure modes in AI clinical trials \cite{liu_consort_ai_2020, rivera_spirit_ai_2020}. But they do not resolve the deeper issue when a model is evaluated against a negotiated, variable ground truth using metrics that average over a heterogeneous population. What does a high-performance figure certify? It certifies that the model agrees with a particular consensus, averaged over a particular sample, under particular acquisition conditions. It does not certify clinical safety or generalizability, a concern that is especially acute in Norway, where digitized bone marrow smear analysis is at an early stage in clinical practice and available training data is limited.

Being right, in this third sense, requires situating performance claims within an honest account of their limits: the specific population over which they were measured, the ground truth against which they were calculated, and the failure modes that aggregate metrics do not surface. This means pairing quantitative evaluation with qualitative clinical validation, prospective deployment studies, and post-market monitoring — all required in principle by the EU AI Act's high-risk framework, but rarely described in adequate detail in published performance studies.

\section{Risk of automation bias}

The final dimension of correctness concerns how model outputs are integrated into clinical workflows and how responsibility is distributed when diagnoses go wrong. This is where the technical and the ethical most directly meet.
Human-in-the-loop design reflects a substantive ethical commitment: that AI in high-stakes diagnostic settings should augment rather than replace human judgment, and that the clinical expert should remain the locus of both decision-making authority and accountability. The WHO principle of human autonomy explicitly requires that AI systems preserve the capacity of clinicians to exercise independent judgment and that 'the final decision should remain with humans'. The EU AI Act's classification of diagnostic AI as high-risk requires ongoing human oversight as a condition of deployment.

These principles are sound. But they are threatened in practice by automation bias, which is the tendency of human operators to over-rely on automated outputs, reducing scrutiny, accepting incorrect recommendations without critical evaluation, and deferring to the system even when independent judgment would lead to a different and correct conclusion [12]. Automation bias is most dangerous in settings where the operator is time-pressured, the task is cognitively demanding, and the automated system is perceived as authoritative. All three conditions characterize hematology practice: pathologists work under significant time pressure, morphological analysis of bone marrow smears is technically demanding, and an AI system that has been validated and approved for clinical use will carry perceived authority.

The irony is that the primary motivation for developing AI-assisted bone marrow analysis is to address a shortage of specialized expertise and to reduce the time burden on pathologists. But as routine aspects of smear analysis are offloaded to the AI, the pathologist's direct engagement with the full morphological spectrum may diminish, as may the perceptual fluency needed to critically evaluate AI outputs and recognize when the system is wrong. Over time, the human nominally in the loop may become progressively less equipped to serve as a genuine check on the system's errors.

Addressing automation bias requires design choices that do not merely include a human in the workflow but structure that involvement to remain active. This means presenting model outputs alongside confidence levels and explanations that demand evaluation rather than passive acceptance; requiring explicit documented confirmation of AI recommendations before they enter the medical record; designing interfaces that make disagreement easy rather than effortful; and building audit mechanisms that allow error patterns to be identified and fed back into model improvement. It also requires institutional commitment to maintaining pathologist expertise in morphological analysis, even as AI handles increasing volume, not because AI cannot handle volume, but because human expertise in the domain is a prerequisite for meaningful oversight.

The question of accountability when AI-assisted diagnosis harms a patient remains philosophically and legally unresolved. The EU AI Act establishes requirements for post-market monitoring and traceability, but it does not resolve how responsibility is distributed in a complex hybrid human-AI decision process. Is the clinician responsible for accepting a recommendation they did not adequately scrutinize? The developer responsible for deploying a model that was overconfident in a domain where overconfidence was foreseeable? Is the institution responsible for implementing the system without adequate safeguards? These questions will require not just legal clarification but new professional norms about what constitutes responsible use of AI assistance in diagnostic medicine — norms that do not yet exist in stable form.

\section{Conclusion}

This paper has argued that correctness in medical AI is not a single property but a set of interdependent requirements, each demanding a different form of honesty and accountability. A model that achieves high accuracy against a carefully constructed test set may nonetheless be wrong in senses that matter more for clinical practice. It may have learned a negotiated, institution-specific approximation of diagnostic truth. It may suppress uncertainty precisely where uncertainty is most clinically warranted. It may perform well on aggregate metrics while failing systematically on the rare cells that determine a diagnosis. It may be deployed in workflows that structurally undermine the human oversight it is designed to support. 

These arguments have been developed through the specific context of plasma cell classification in bone marrow smears for multiple myeloma, but their implications extend broadly across medical AI. The WHO ethics framework, the EU AI Act, and clinical reporting guidelines like CONSORT-AI and DECIDE-AI all represent important progress toward embedding ethical requirements in the design, evaluation, and governance of medical AI systems.

\bibliographystyle{unsrt}
\bibliography{ref}

@article{liu_consort_ai_2020,
  author  = {Liu, Xiaoxuan and Rivera, Steven C. and Moher, David and Calvert, Melanie J. and Denniston, Alastair K. and SPIRIT-AI and CONSORT-AI Working Group},
  title   = {Reporting guidelines for clinical trial reports for interventions involving artificial intelligence: the CONSORT-AI extension},
  journal = {Nature Medicine},
  volume  = {26},
  number  = {9},
  pages   = {1364--1374},
  year    = {2020},
  doi     = {10.1038/s41591-020-1034-x}
}

@article{rivera_spirit_ai_2020,
  author  = {Rivera, Steven C. and Liu, Xiaoxuan and Chan, Adrian W. and Denniston, Alastair K. and Calvert, Melanie J. and SPIRIT-AI and CONSORT-AI Working Group},
  title   = {Guidelines for clinical trial protocols for interventions involving artificial intelligence: the SPIRIT-AI extension},
  journal = {Nature Medicine},
  volume  = {26},
  number  = {9},
  pages   = {1351--1363},
  year    = {2020},
  doi     = {10.1038/s41591-020-1037-7}
}

@inproceedings{guo2017calibration,
  title     = {On Calibration of Modern Neural Networks},
  author    = {Guo, Chuan and Pleiss, Geoff and Sun, Yu and Weinberger, Kilian Q.},
  booktitle = {Proceedings of the 34th International Conference on Machine Learning (ICML)},
  pages     = {1321--1330},
  year      = {2017},
  organization = {PMLR}
}

@book{who2024ai,
  title     = {Ethics and governance of artificial intelligence for health: Guidance on large multi-modal models},
  author    = {{World Health Organization}},
  year      = {2024},
  publisher = {World Health Organization},
  address   = {Geneva}
}

@article{decideai2022,
  title   = {Reporting guideline for the early-stage clinical evaluation of decision support systems driven by artificial intelligence: DECIDE-AI},
  author  = {{DECIDE-AI Expert Group}},
  journal = {Nature Medicine},
  volume  = {28},
  number  = {5},
  pages   = {924--933},
  year    = {2022},
  doi     = {10.1038/s41591-022-01772-9}
}

@book{popper1959logic,
  title     = {The Logic of Scientific Discovery},
  author    = {Popper, Karl R.},
  year      = {1959},
  publisher = {Hutchinson},
  address   = {London}
}

@article{northcutt2021confident,
  title   = {Confident Learning: Estimating Uncertainty in Dataset Labels},
  author  = {Northcutt, Curtis G. and Jiang, Lu and Chuang, Isaac L.},
  journal = {Journal of Artificial Intelligence Research},
  volume  = {70},
  pages   = {1373--1411},
  year    = {2021}
}

@article{elmore2015diagnostic,
  title   = {Diagnostic concordance among pathologists interpreting breast biopsy specimens},
  author  = {Elmore, Joann G. and Longton, Gary M. and Carney, Patricia A. and others},
  journal = {JAMA},
  volume  = {313},
  number  = {11},
  pages   = {1122--1132},
  year    = {2015}
}

@inproceedings{nixon2019measuring,
  title     = {Measuring Calibration in Deep Learning},
  author    = {Nixon, Jeremy and Dusenberry, Michael W. and Zhang, Linchuan and Jerfel, Ghassen and Tran, Dustin},
  booktitle = {Proceedings of the IEEE/CVF Conference on Computer Vision and Pattern Recognition Workshops (CVPRW)},
  year      = {2019}
}

@article{secondopinion2021,
  title   = {Second opinion needed: communicating uncertainty in medical machine learning},
  author  = {Kelly, Christopher J. and Karthikesalingam, Alan and Suleyman, Mustafa and Corrado, Greg and King, Dominic},
  journal = {Nature Medicine},
  volume  = {27},
  pages   = {203--209},
  year    = {2021},
  doi     = {10.1038/s41591-020-01179-0}
}

@incollection{li2022who5th,
  author       = {Li, Weijie},
  title        = {The 5th Edition of the World Health Organization Classification of Hematolymphoid Tumors},
  booktitle    = {Leukemia [Internet]},
  publisher    = {StatPearls Publishing / NCBI Bookshelf},
  year         = {2022},
  month        = {August},
  url          = {https://www.ncbi.nlm.nih.gov/books/},
  note         = {Available from: NCBI Bookshelf},
}

@book{fleck1979genesis,
  author    = {Fleck, Ludwik},
  title     = {Genesis and Development of a Scientific Fact},
  publisher = {University of Chicago Press},
  year      = {1979},
  note      = {Originally published in 1935}
}

@misc{eu2024aiact,
  author       = {{European Parliament and Council of the European Union}},
  title        = {Regulation (EU) 2024/1689 of the European Parliament and of the Council of 13 June 2024 laying down harmonised rules on artificial intelligence (Artificial Intelligence Act)},
  year         = {2024},
  howpublished = {\url{https://eur-lex.europa.eu/eli/reg/2024/1689/oj}},
  note         = {Official Journal of the European Union, L 2024/1689, 12 July 2024}
}
\end{document}